\documentclass[10pt,twocolumn,letterpaper]{article}
\usepackage[dvipsnames]{xcolor}
\usepackage[pagenumbers]{cvpr} % To force page numbers, e.g. for an arXiv version

\def\paperID{2397} % *** Enter the Paper ID here
\def\confName{CVPR}
\def\confYear{2025}

\usepackage{multicol}
\usepackage[dvipsnames]{xcolor}
\usepackage[T1]{fontenc}
\usepackage[utf8]{inputenc}
\usepackage{figchild}
\usepackage{tikz}
\usepackage{tikzsymbols}
\usepackage{epsfig}
\usepackage{graphicx}
\usepackage{amsmath}
\usepackage{amssymb}
\usepackage{float}
\usepackage{booktabs} % for professional tables
\usepackage{mathtools}
\usepackage{amsthm}
\RequirePackage{algorithm, algorithmic}
\usepackage{multirow}
\usepackage{caption}
\usepackage{utfsym}
\usepackage[pagebackref,breaklinks,colorlinks]{hyperref}
\theoremstyle{plain}

\theoremstyle{definition}

\theoremstyle{remark}

\definecolor{myblue}{HTML}{5364cc}

\usepackage{array} 
\usepackage{fontawesome5}
\usepackage{figchild}
\usepackage{utfsym}

\newif\ifdraft
\drafttrue
\ifdraft
     \newcommand{\wm}[1]{\textcolor{magenta}{{[wwm: #1]}}}
\else
    \newcommand{\wm}[1]{}
\fi

\definecolor{deepgreen}{rgb}{0,0.5,0.4}
\definecolor{teal}{rgb}{0.78, 0.875, 0.703}

\usepackage{pifont}% http://ctan.org/pkg/pifont
\title{\textcolor{myblue}{NoPain}: \textcolor{myblue}{N}o-b\textcolor{myblue}{o}x \textcolor{myblue}{P}oint Cloud \textcolor{myblue}{A}ttack via Optimal Transport S\textcolor{myblue}{i}ngular Bou\textcolor{myblue}{n}dary}

\author{Zezeng Li$^{1,2}$, ~ Xiaoyu Du$^{1}$, ~Na Lei$^{1*}$
, ~Liming Chen$^{2,3,4}$, ~Weimin Wang$^1$\thanks{Corresponding authors.
}\\ 	
\normalsize{$^1$School of Software, Dalian University of Technology, China}\qquad
\normalsize{$^2$École Centrale de Lyon, France}\\
\normalsize{$^3$University of Lyon, France}
\qquad
\normalsize{$^4$Institut Universitaire de France}
}

\begin{document}
\maketitle

%%%%%%%%% ABSTRACT
\begin{abstract} 
Adversarial attacks exploit the vulnerability of deep models against adversarial samples. Existing point cloud attackers are tailored to specific models, iteratively optimizing perturbations based on gradients in either a white-box or black-box setting. Despite their promising attack performance, they often struggle to produce transferable adversarial samples due to overfitting to the specific parameters of surrogate models. To overcome this issue, we shift our focus to the data distribution itself and introduce a novel approach named \textbf{NoPain}, which employs optimal transport (OT) to identify the inherent singular boundaries of the data manifold for cross-network point cloud attacks. Specifically, we first calculate the OT mapping from noise to the target feature space, then identify singular boundaries by locating non-differentiable positions. Finally, we sample along singular boundaries to generate adversarial point clouds. Once the singular boundaries are determined, NoPain can efficiently produce adversarial samples without the need of iterative updates or guidance from the surrogate classifiers. Extensive experiments demonstrate that the proposed end-to-end method outperforms baseline approaches in terms of both transferability and efficiency, while also maintaining notable advantages even against defense strategies. Code and model are available at \url{https://github.com/cognaclee/nopain}.
\end{abstract}

%%%%%%%%% BODY TEXT
\vspace{-2mm}
\section{Introduction}
Recent research has extensively examined the adversarial vulnerability of deep neural networks (DNNs) \cite{szegedy2013intriguing,goodfellow2014FGSM,moosavi2016deepfool,papernot2016limitations,carlini2017cw,athalye2018obfuscated,duboudin2022look}, demonstrating that even minimal perturbations to input data can lead advanced DNN models to make erroneous predictions. This vulnerability poses significant threats to security-critical systems and has spurred research on adversarial attacks to improve models' robustness. 

\begin{figure}[htb!]
\centering
\includegraphics[width=.99\linewidth]{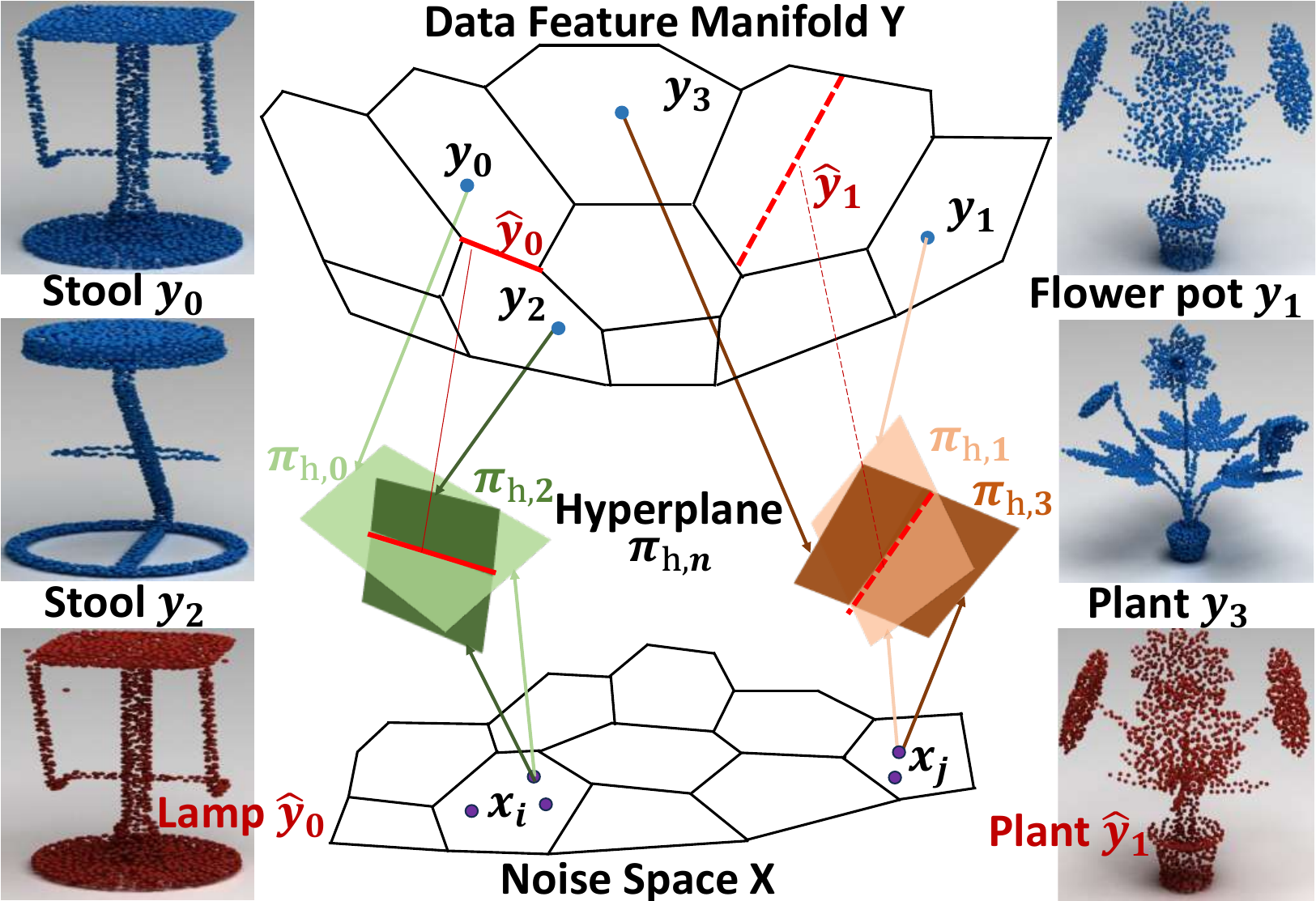}
\vspace{-2mm}
\caption{Point cloud attack via OT singular boundary. We exploit singular boundaries of the data manifold, induced by the OT, to perform no-box attacks. Our approach begins by applying the OT mapping to obtain the hyperplane
set \{\textcolor{orange}{$\pi_{\boldsymbol{h},i}$}\} and polygons decomposition of the data manifold. Next, we compute dihedral angles between neighbor hyperplanes to identify singular boundaries. Finally, adversarial samples \textcolor{purple}{$\boldsymbol{\widehat{y}}$} are generated by sampling along singular boundaries. 
Hyperplanes of the same color represent the hyperplane associated with $\boldsymbol{y_i}$~(\textcolor{deepgreen}{dark}) and its neighbor (\textcolor{teal}{light}), with a singular boundary indicated by the \textcolor{red}{solid and dashed red lines}.}
\label{fig:illustration}
\end{figure}

Given the crucial role of 3D DNNs in security-sensitive applications such as autonomous driving and robot navigation, various point cloud attack methods~\cite{liu2019extending,zhang2019defense,tsai2020robust,hamdi2020advpc,sun2021adversarially,liu2020sink,rampini2021universal,cheng2021universal,dong2022isometric,huang2022siadv,liu2022boosting,tao20233dhacker,tang2023manifold,zhang2024improving} have been developed to perturb data from different perspectives, effectively revealing the vulnerabilities of current point cloud classifiers. Most of these methods are white-box attacks~\cite{liu2019extending,zhang2019defense,xiang2019generating,tsai2020robust,sun2021adversarially,kim2021minimal,rampini2021universal,cheng2021universal,shi2022shape,dong2022isometric,tang2023manifold,lou2024hide}, requiring access to the structure, weights, and gradients of the target models. However, their effectiveness diminishes significantly when tested on different networks, indicating low attack transferability. %Its cause lies in the tendency for optimization-based attacks to overfit specific network parameters in a white-box setting.

Contemporary advancements aimed at enhancing attack transferability in black-box settings~\cite{hamdi2020advpc,liu2022boosting,huang2022siadv,liu2022imperceptible,zhang2024eidos,tao20233dhacker,he2023generating,chen2024anf,zhang2024improving} can be divided into two primary categories: transfer-based attacks and boundary-based attacks. Transfer-based methods typically employ autoencoders or partial parameters of the surrogate model to enhance the transferability of attacks. Boundary-based attacks aim to improve transferability by generating perturbations at the decision boundaries, thereby altering the predictions. While these techniques enhance transferability, they require access to partial model parameters or multiple queries to proxy models for iterative optimization of adversarial samples. This reliance on model-specific strategies introduces the risk of overfitting, ultimately limiting their transferability.

Some researchers have approached attack by focusing on global distribution alignment. As a powerful tool for distribution alignment, optimal transport~(OT) has been successfully applied to transferable attacks in images \cite{han2023ot,labarbarie2024optimal,li2023ot}. Han \etal~\cite{han2023ot} leverage OT to align image and text distributions, enhancing the transferability of attacks in image-language models. Labarbarie \etal~\cite{labarbarie2024optimal} achieve a patch adversarial attack by aligning features of adversarial images produced by the surrogate classifiers' encoder with those of target images. This raises key questions: \textbf{Is a surrogate model necessary? Must adversarial samples be obtained through optimization?} In response, this paper explores a no-box~(classifier-free) end-to-end point cloud attack.

To achieve this, we first approach adversarial attacks as a generative task. By calculating the OT mapping from the noise space to the feature space, we identify the local singular boundaries of the data manifold, represented by feature pairs where the OT mapping is non-differentiable. We then perturb features by shifting them toward these boundaries, generating modified features. Building on it, we introduce \textbf{NoPain}, a method capable of generating highly transferable examples without iterative optimization or supervision from surrogate models. In summary, our main contributions are:
\begin{itemize}[leftmargin=*]
\item We propose a novel no-box adversarial attack framework by directly exploring the data manifold's singular boundaries with explicit and geometric interpretable OT map.
\item Our algorithm exhibits strong cross-network transferability and robustness against defense owing to being free from model-specific loss and leveraging intrinsic data characteristics.
\item The proposed method is end-to-end and requires no optimization, significantly enhancing the attack's efficiency.
\item Extensive experiments show that \textbf{NoPain} outperforms the SOTAs regarding the attack performance and adversary quality, particularly in transferability.
\end{itemize}

\begin{figure*}[tbhp!]
\vspace{-2mm}
  \centering
   \includegraphics[width=0.99\linewidth]{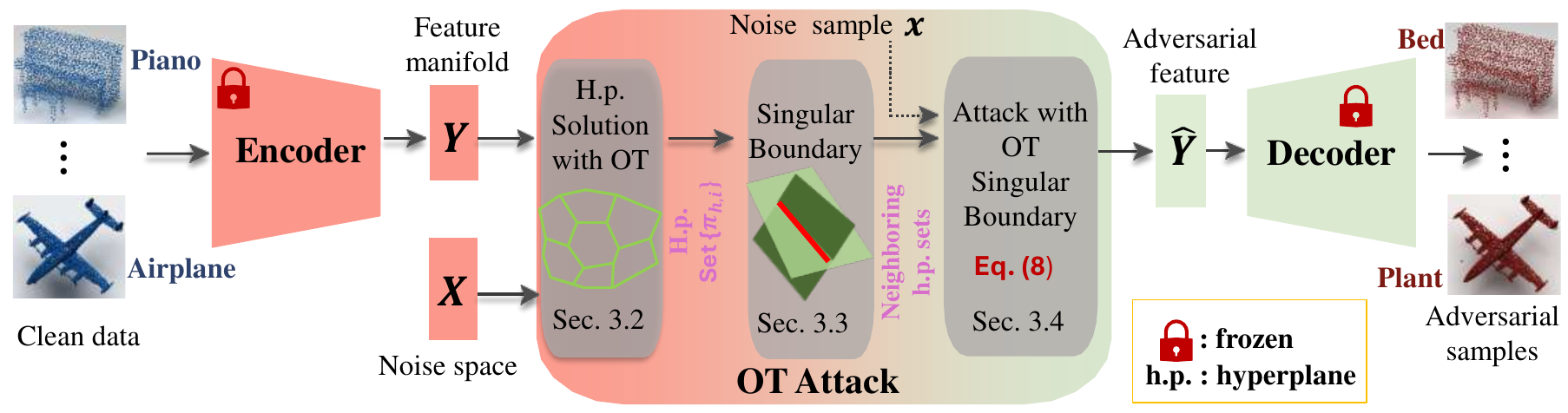}
   \vspace{-3mm}
   \caption{Overview of the proposed no-box point cloud attack framework \textbf{NoPain}. $\boldsymbol{Y}$ represents sample features, and $\boldsymbol{X}$ is noise. The dotted line indicates the process only in the test phase. The blue point cloud on the left is the original point cloud, and the crimson one on the right represents the generated adversarial samples. 
   For the OT Attack, we first apply OT to calculate the hyperplane set, {$\pi_{\boldsymbol{h},i}$}, associated with each feature $\boldsymbol{y}_{i}$. Next, we use the approach in Sec.\ref{sec:singbound} to determine singular boundaries and execute the attack with Eq.\eqref{eq:interplatation} in Sec.~\ref{sec:sampling}.}
   \label{fig:framework}
   \vspace{-2mm}
\end{figure*}

\section{Related Work}
\noindent \textbf{White-box attack on point cloud.}
Existing works of point cloud attacks can be roughly divided into white-box attacks and black-box attacks. %White box attack means that all information of the attacked classification model can be used to update perturbations, including the prediction and gradient of the model. 
Recently, most white-box attack works adopt \emph{point-based attacks}~\cite{liu2019extending,zhang2019defense,tsai2020robust,sun2021adversarially}. Liu et al.~\cite{liu2019extending} extended the gradient-based adversarial attack FGSM~\cite{goodfellow2014FGSM} strategy to point clouds.%, iteratively searching the desired point-wise perturbation under an $l_2$ norm constraint.
3D-Adv~\cite{xiang2019generating} introduced the C\&W attack framework~\cite{carlini2017cw} in point cloud attacks, producing adversarial examples by shifting point coordinates and adding additional points.
%with the guidance of the target classifier. Afterward, 
Tsai \etal~\cite{tsai2020robust} improved the C\&W attack framework by introducing a KNN regularization term to suppress outlier points and compact point cloud surface. GeoA\text{$^3$}~\cite{wen2020geometry} uses a combined geometry-aware objective to maintain local curvature consistency and a uniform surface.% on the adversarial point cloud. 
Zheng \etal~\cite{zheng2019saliency} proposed that deleting a small number of points with high saliency can effectively cause misclassification.
Apart from the point-based methods mentioned above, several studies have explored \emph{shape-based attack}. Liu \etal~\cite{liu2020sink} introduced shape-based attacks by adding new features to objects. %, as well as using a Gaussian kernel for sinking operations to deform point clouds. 
Zhang \etal~\cite{zhang2023meshattack} and Miao \etal~\cite{dong2022isometric} proposed directly attacking the mesh to generate smoother results.%, employing edge length regularization and Gaussian curvature regularization, respectively.
Tang \etal~\cite{tang2023manifold} proposed to adversarially stretch the latent variables in an auto-encoder, which can be decoded as smooth adversarial point clouds. %Lou \etal~\cite{lou2024hide} proposed a shape-based attack HiT-ADV, which conducts a two-stage search for attack regions based on saliency and imperceptibility scores, and then adds deformation perturbations in each attack region using Gaussian kernel functions. 
HiT-ADV~\cite{lou2024hide} is a shape-based attack, that first search attack regions based on saliency and imperceptibility scores, and then adds deformation perturbations in each attack region with Gaussian kernel. 
Besides, some works \cite{kim2021minimal,shi2022shape,dong2022isometric,tang2023manifold} attack point clouds in the feature space for imperceptible attack. While these methods eliminate outliers and ensure smoothness, they still require optimization for each point cloud, resulting in a high time cost.  To this end, universal attack~\cite{rampini2021universal,cheng2021universal} was proposed to compute universal perturbations for point clouds with specific patterns.

\noindent \textbf{Black-box attack on point cloud.}
The black-box attack can be further classified as transfer-based~\cite{hamdi2020advpc,liu2022imperceptible,huang2022siadv,liu2022boosting,zhang2024eidos,zhang2024improving,cai2024frequency} and boundary-based black-box attacks~\cite{tao20233dhacker,he2023generating,chen2024anf}. For point cloud, most works focus on transfer-based black-box attacks. AdvPC~\cite{hamdi2020advpc} leverages a point cloud auto-encoder to enhance the transferability of adversarial point clouds, while AOF~\cite{liu2022boosting} targets the low-frequency components of 3D point clouds to disrupt general features. SI-Adv~\cite{huang2022siadv} projects points onto a tangent plane and introduces perturbations to create shape-invariant point clouds. Eidos~\cite{zhang2024eidos} is a transfer-based attack that allows adversarial examples trained on one classifier to be transferred to another. 3DHacker~\cite{tao20233dhacker} generates adversarial samples using only black-box hard labels. PF-Attack~\cite{he2023generating} and ANF~\cite{chen2024anf} optimize perturbations and their subcomponents through adversarial noise factorization near decision boundaries, reducing dependency on surrogate models and enhancing transferability. SS-attack~\cite{zhang2024improving} applies random scaling or shearing to the input point cloud to prevent overfitting the white-box model, thus improving attack transferability. While these methods enhance model transferability, they still rely on iterative label-based generation of adversarial samples.

\noindent\textbf{No-box attacks.} The no-box approach is a classifier-free attack strategy that requires neither access to classifier details nor model queries. To date, only a few studies have addressed this challenging setup for images or skeletons. Li \etal~\cite{li2020practical} employed an autoencoding model to design an adversarial loss for no-box image attacks. Sun \etal~\cite{sun2022towards} used a small subset of the training set to train an auxiliary model, leveraging this model to generate adversarial examples and attack the target model. Lu \etal~\cite{lu2023hard} define an adversarial loss to maximize each adversary's dissimilarity with positive samples while minimizing its similarity with negative samples for the skeleton attack.
Zhang \etal~\cite{zhang2022practical} combined the low frequency of a clean image with the high frequency of a texture image to craft adversarial examples. Mou \etal~\cite{mou2024no} developed a decision-based attack strategy that generates universal adversarial perturbations and a set of texture-adversarial instances.

\vspace{-1mm}
\noindent \textbf{Boundary-based attacks.}
Boundary-based attack method \cite{brendel2018decision} is widely used in the 2D field, which is an efficient framework that uses the final decision results to implement black-box attacks. In the 2D field, the decision boundary attack process starts with two origin images called 
source-image and target-image with different labels. Then, it performs a binary search to obtain a boundary image on the decision boundary. 
Various 2D decision boundary-based attacks are proposed based on this general attack framework. Thomas \etal~\cite{brunner2019guessing} and Vignesh \etal~\cite{srinivasan2019black} propose to
choose more efficient random perturbation including Perlin noise and DCT in random walking steps instead of Gaussian perturbation. Chen \etal~\cite{chen2020hopskipjumpattack} conduct a gradient estimation method using the Monte-Carlo sampling strategy instead of random perturbation. Thereafter, several works~\cite{li2020qeba,li2021nonlinear,li2022decision} improve the gradient estimation strategy through sampling from representative low-dimensional subspace. Recently, 
Tao \etal~\cite{tao20233dhacker} introduced boundary-based black-box attacks on point clouds, proposing 3DHacker, which leverages a developed decision boundary algorithm to attack point clouds using only black-box hard labels. He \etal~\cite{he2023generating} and Chen \etal~\cite{chen2024anf} jointly optimize two sub-perturbations near decision boundaries via adversarial noise factorization, enhancing transferability. %However, these three boundary-based point cloud attack methods are all optimization-based, requiring optimization for each adversarial sample with guidances of specific models, leading to a higher time cost and imperfect transferability.
However, these boundary-based point cloud attack methods require optimization for each adversarial sample using model-specific guidance, resulting in higher time costs and limited transferability.

\section{Methodology}
Here, we are committed to providing a no-box end-to-end point cloud attack by incorporating the optimal transport singular boundary. As illustrated in \cref{fig:framework}, our framework \textbf{NoPain} comprises three stages. Firstly, the input point clouds are embedded into the latent space, obtaining the feature vectors. Secondly, we obtain the singular boundaries of the point cloud data manifold by solving the OT mapping from noise to the features space, and then perturb the features by shifting them toward these boundaries. Finally, a pre-trained decoder was utilized to generate transferable adversarial point clouds in an end-to-end fashion. 

\noindent\textbf{Motivations.} 
Although existing point cloud attack methods have demonstrated high attack quality, they struggle to produce transferable adversarial samples~\cite{cai2024frequency,zhang2024improving,he2023generating}.%, as depicted in \cref{fig:qualitative}. 
This limitation stems from the tendency of optimization-based attacks to overfit specific parameters of surrogate networks. To address this, we shift our focus to the data itself, aiming to uncover the inherent characteristics of target data distribution. Notably, considering mode mixture at the singular boundary, we propose leveraging singular boundaries to achieve cross-network point cloud attacks.

Compared to existing point cloud attack methods, our approach offers several key advantages and fundamental distinctions: 1)~It eliminates the need for surrogate classifiers; 2)~It conducts attacks based on optimal transport singular boundaries, providing greater interpretability due to the explicit solution of the OT mapping; 3)~It operates without iterative optimization. %While traditional methods often target overfitting or insufficient generalization of classification models, our approach directly exploits the intrinsic characteristics of the data distribution.  
By leveraging the intrinsic characteristics of the data distribution, our method achieves no-box, end-to-end transferable point cloud attacks.

\subsection{Problem formulation}
\label{subsec:preliminary}
Given a point cloud dataset $\boldsymbol{\mathcal{P}} =\left\{\boldsymbol{P}_i\right\}_{i=1}^{N}$ with $N$ point cloud, our goal is to generate a set of adversarial point clouds $\boldsymbol{\mathcal{\widehat{P}}} =\left\{\boldsymbol{\widehat{P}}_i\right\}_{i=1}^{N}$ with sufficiently small perturbations (i.e., small $||\boldsymbol{\widehat{P}}_i - \boldsymbol{P}_i||$) such that $f(\boldsymbol{\widehat{P}}_i) \neq f(\boldsymbol{P}_i)$ for all $\boldsymbol{P}_i\in\boldsymbol{\mathcal{P}}$, where $f$ is a unknown classifier during the attack.

To achieve this, we first embed the point cloud into a hidden space manifold using an encoder, $\mathbf{E}\phi$, resulting in the feature representation $\boldsymbol{y} = \mathbf{E}\phi(\boldsymbol{P})$. We then detect local singular boundaries on the target data manifold and launch attacks based on these boundaries. To identify these singular boundaries, we solve a semi-discrete OT mapping from a continuous noise space to discrete data points, forming a hyperplane defined by noise $\boldsymbol{X}$ and target data $\boldsymbol{Y}$. These hyperplanes enable us to determine singular boundaries within the data manifold effectively. In the following, we will introduce relevant OT theories to provide the foundation for subsequent method explanations.

\noindent\textbf{Semi-discrete Optimal Transport} 
Suppose the source measure $\mu$ defined on a convex domain $\Omega \subset \mathbb{R}^{d}$, the target domain is a discrete set $\boldsymbol{Y}=\left\{\boldsymbol{y}_{i}\right\}_{i=1}^{N}, \boldsymbol{y}_{i} \in \mathbb{R}^{d}$. The target measure is a Dirac measure $\nu=\sum_{i=1}^{N} \nu_{i} \delta\left(\boldsymbol{y}-\boldsymbol{y}_{i}\right)$ and the source measure is equal to total mass as $\mu(\Omega)=\sum_{i=1}^{N} \nu_{i} .$ Under a semi-discrete transport mapping $g: \Omega \rightarrow \boldsymbol{Y}$, a cell decomposition is induced $\Omega=\bigcup_{i=1}^{N} W_{i}$, such that every $\boldsymbol{x}$ in each cell $W_{i}$ is mapped to the target $\boldsymbol{y}_{i}, g: \boldsymbol{x} \in W_{i} \mapsto \boldsymbol{y}_{i}$. The mapping $g$ is measure preserving, denoted as $g_{\#} \mu=\nu$, if the $\mu$-volume of each cell $W_{i}$ equals to the $\nu$-measure of the image $g\left(W_{i}\right)=\boldsymbol{y}_{i}, \mu\left(W_{i}\right)=\nu_{i}$. The cost function is given by $c: \Omega \times \boldsymbol{Y} \rightarrow \mathbb{R}$, where $c(\boldsymbol{x}, \boldsymbol{y})$ represents the cost for transporting a unit mass from $\boldsymbol{x}$ to $\boldsymbol{y}$. The semi-discrete OT~(SDOT) mapping $g^{\ast}$ is a measure-preserving mapping that minimizes the total cost in Eq.~\eqref{eq:SDOT},
\begin{equation}
\vspace{-2mm}
g^{\ast}:=\arg \min _{g_{\#} \mu=\nu}\sum_{i=1}^{N} \int_{W_{i}} c(\boldsymbol{x}, g(\boldsymbol{x})) d \mu(\boldsymbol{x}).\label{eq:SDOT}
\end{equation}
\vspace{-2mm}

According to Brenier theorem~\cite{Brenier1991}, when the cost function $c(\boldsymbol{x}, \boldsymbol{y})=1 / 2\|\boldsymbol{x}-\boldsymbol{y}\|^{2}$, we have
$g^{*}(\boldsymbol{x})=\nabla \boldsymbol{u}(\boldsymbol{x}).$
This explains that the SDOT mapping is the gradient mapping of Brenier's potential $\boldsymbol{u}$. As~\cite{lei2020geometric,an2019ae} remark, $\boldsymbol{u}$ is the upper envelope of a collection of hyperplanes
\begin{equation}\label{eq:hyperplanes}
\pi_{\boldsymbol{h},i}(\boldsymbol{x}) =
% \boldsymbol{x}^T\boldsymbol{y}_i+h_i\, 
\langle\boldsymbol{y}_i, \boldsymbol{x} \rangle+h_i\,.
\end{equation}
Specifically, $\boldsymbol{u}$ can be parametrized uniquely up to an additive constant by the Brenier's height vector $\boldsymbol{h} = (h_1, h_2, ..., h_{N})^T$ and can be stated as follows,
\begin{equation}
\boldsymbol{u}_{\boldsymbol{h}}(\boldsymbol{x}) = \max_{i=1}^{N}\{\pi_{\boldsymbol{h},i}(\boldsymbol{x})\}, \boldsymbol{u}_{\boldsymbol{h}}: \Omega \rightarrow \mathbb{R}^n,\label{eq:uh}
\end{equation}

The way in which Brenier's potential $\boldsymbol{u}_{\boldsymbol{h}}$ maximizes the hyperplane induces the cell decomposition $\Omega=\bigcup_{i=1}^{N} W_{i}$ for $\boldsymbol{X}$, and also implicitly establishes the polygons decomposition on the target domain $\boldsymbol{Y}$. The edges of each polygon represent the boundaries between hyperplanes. Thus, the pertinent issue that needs to be considered next is how to solve the hyperplane set, i.e. the height vector $\boldsymbol{h}$.

\subsection{Hyperplane Set Solution}
\begin{algorithm}[tb]
   \caption{OT Solver for Hyperplane Set}
   \label{alg:OTMap}
\begin{algorithmic}[1]
   \REQUIRE  Dataset $\boldsymbol{Y}=\left\{\boldsymbol{y}_i\right\}_{i=1}^{N}$, initial noise sample number $M$, learning rate $lr$, threshold $\eta$, positive integer $s$.
   \STATE Initialize $\boldsymbol{h}=(h_1,h_2,\cdots,h_{N})\leftarrow(0,0,\cdots,0)$.
   \REPEAT
    \STATE Sample $M$ noise samples $\boldsymbol{X} = \left\{\boldsymbol{x}_j\sim \mathcal{N}(0,I)\right\}_{j=1}^M$
    \STATE $w(\boldsymbol{h}) = (0,0,\cdots,0)$. 
    %\STATE $\pi_{\boldsymbol{h},i}(\boldsymbol{x})$ is calculated by Eq.\eqref{eq:hyperplanes} $\forall$ $\boldsymbol{x}_j$, $\boldsymbol{{y}_i}\in\boldsymbol{Y}$.
    \FOR{ $j=0;j < M$} 
    %\STATE  Calculate the index of the maximum $\pi_{\boldsymbol{h},i}(\boldsymbol{x}_j)$ for $\boldsymbol{x}_j$, $index= k$ if $\max\{\pi_{\boldsymbol{h},i}\}_{i=1}^{N}==\pi_{\boldsymbol{h},k}$.
    \STATE  $k=\arg\max _{i\in\{1,\cdots,N\}} \pi_{\boldsymbol{h},i}(\boldsymbol{x}_j)$ with Eq.\eqref{eq:hyperplanes}.
    % \STATE $w_k(\boldsymbol{h})\leftarrow w_k(\boldsymbol{h})+1$
    \STATE $w(\boldsymbol{h})[k]\leftarrow w(\boldsymbol{h})[k]+1$. \COMMENT{\textcolor{blue}{[k] indicates the indexing operation}}
    \STATE  $j\leftarrow j+1$.
    \ENDFOR
    \STATE $w(\boldsymbol{h})\leftarrow \frac{w(\boldsymbol{h})}{M}$.
    \STATE Calculate $\nabla \boldsymbol{h}\leftarrow(w(\boldsymbol{h})-\frac{1}{N} )^T$.
    \STATE $\nabla \boldsymbol{h}\leftarrow\nabla \boldsymbol{h}-\text{mean}(\nabla \boldsymbol{h})$.
    \STATE Update $\boldsymbol{h}$ by Adam algorithm.% with $\beta_1=0.9$,$\beta_2=0.5$.
    %\STATE Calculate $ E(\boldsymbol{h})$ by Eq.~\eqref{eq:EH}
    \IF{$ E(\boldsymbol{h})$ in Eq.~\eqref{eq:EH} has not decreased for $s$ steps}
    \STATE $M\leftarrow 2\times M$; $lr\leftarrow 0.8\times lr$\,.
    \ENDIF
   \UNTIL{$E(\boldsymbol{h})<\eta$}
   %\STATE OT mapping $g(\cdot)\leftarrow \nabla(\max_{i}\{\langle \boldsymbol{x}_T,\boldsymbol{y}_i\rangle_F  + h_i\})$.
   \STATE {\bf Return} Brenier's height vector $\boldsymbol{h} = (h_1, h_2, ..., h_{N})$.
\end{algorithmic}
\end{algorithm}
\noindent Given the Target dataset $\boldsymbol{Y}=\left\{\boldsymbol{y}_i\right\}_{i=1}^{N}$ with target measure $\nu$, there exists Brenier's potential $\boldsymbol{u}_{\boldsymbol{h}}$ in Eq.~\eqref{eq:uh} whose projected volume of each support plane is equal to the given target measure $\nu_i$\cite{Brenier1991,an2019ae}. 
To obtain adversarial samples for all point clouds in the dataset, we set the target measure to a uniform distribution, i.e. $\nu_i=\frac{1}{N},\ \forall i=1,\cdots,N$. Then, we can get the optimal $\boldsymbol{h}$ and $\boldsymbol{u}_{\boldsymbol{h}}$ by minimizing the following convex energy function:%\sum_{i=1}^{N}
\begin{equation}
E(\boldsymbol{h})=\sum_{i=1}^{N}( w_{i}(\boldsymbol{h}) -\frac{1}{N})^2,
\label{eq:EH}
\end{equation}
where $\omega_{i}(\boldsymbol{h})$ is the $\mu$-volume of $W_{i}(\boldsymbol{h})$, i.e., the frequency of $\boldsymbol{x}$ assigned to $\boldsymbol{y}_i$. The energy $E(\boldsymbol{h})$ provides the optimization direction for $\boldsymbol{h}$, and its gradient $\nabla \boldsymbol{h}$ is given by
\begin{equation}
\nabla \boldsymbol{h}=(w(\boldsymbol{h})-\frac{1}{N} )^T\,.
\label{eq:gradient}
\end{equation}
Then, we optimize $\boldsymbol{h}$ using the Adam optimization algorithm~\cite{adam2014}. To ensure a unique solution, we adjust $\nabla \boldsymbol{h}$ to have zero mean by setting $\nabla \boldsymbol{h} = \nabla \boldsymbol{h} - \text{mean}(\nabla \boldsymbol{h})$. 

After obtaining $\boldsymbol{h}$, we directly substitute it into Eq.~\eqref{eq:hyperplanes} to obtain the hyperplane set $\{\pi_{\boldsymbol{h},i}(\boldsymbol{x})|\pi_{\boldsymbol{h},i}(\boldsymbol{x}) =\langle\boldsymbol{y}_{i}, \boldsymbol{x}\rangle+h_i,$ $i=1,\cdots,N\}$. The algorithm is detailed in Algorithm~\ref{alg:OTMap}.

\begin{algorithm}[tb]
   \caption{Point Cloud Attack: \textbf{NoPain}}
   \label{alg:ot-attack}
    \begin{algorithmic}[1]
   \REQUIRE Target dataset $\boldsymbol{\mathcal{P}} =\left\{\boldsymbol{P}_i\right\}_{i=1}^{N}$, a well-trained encoder $\mathbf{E}_\phi$ and decoder $\mathbf{D}_\varphi$, the number of neighbors $K$ in Eq.~\eqref{eq:theta}, threshold $\tau$.
   \ENSURE Generated adversarial samples $\boldsymbol{\mathcal{\widehat{P}}} =\left\{\boldsymbol{\widehat{P}}_i\right\}_{i=1}^{N}$.
   \STATE Embedding $\boldsymbol{\mathcal{P}}$ into latent space with encoders $\mathbf{E}_\phi$, $\boldsymbol{Y}=\left\{\boldsymbol{y}_i|\boldsymbol{y}_i=\mathbf{E}_\phi(\boldsymbol{P}_i),\  i=1,2,\cdots,N\right\}$.
   \STATE The Brenier's height vector $\boldsymbol{h} = (h_1, h_2, ..., h_{N})$ obtained by Algorithm~\ref{alg:OTMap}.
   \STATE Sample $M$ noise samples $\left\{\boldsymbol{x}_j\sim \mathcal{N}(0,I)\right\}_{j=1}^N$.
   \STATE Calculate the hyperplane set $\left\{\boldsymbol{\pi}_{i,j}|\boldsymbol{\pi}_{i,j}=\boldsymbol{x}_j^T\boldsymbol{y}_i+h_i\right\}$ by Eq.~\eqref{eq:hyperplanes}.
   \STATE Calculate dihedral angles $\boldsymbol{\Theta}=\left\{\theta_{i,k}\right\}$ between hyperplanes by Eq.~\eqref{eq:theta}.
   \STATE Obtain point pairs $\left\{(\boldsymbol{y}_{i_0},\boldsymbol{y}_{i_k})\right\}_{i_0=1}^N$ by checking $\boldsymbol{\Theta}$ with threshold $\tau$.
   \STATE Calculate adversarial features $\boldsymbol{\widehat{Y}}=\left\{\boldsymbol{\widehat{y}}_i\right\}_{i=1}^{N}$ by Eq.~\eqref{eq:interplatation}
   \STATE Decode features $\boldsymbol{\widehat{Y}}$ to obtain adversarial samples $\boldsymbol{\mathcal{\widehat{P}}} =\left\{\boldsymbol{\widehat{P}}_i|\boldsymbol{\widehat{P}}_i=\mathbf{D}_{\varphi}(\boldsymbol{\widehat{y}}_i),\  i=1,2,\cdots,N\right\}$.
   \STATE {\bf Return} $\boldsymbol{\mathcal{\widehat{P}}}$
\end{algorithmic}
\end{algorithm}

\vspace{-2mm}
\subsection{Singular Boundary Determination} \label{sec:singbound}
According to Figalli's theory~\cite{figalli2010regularity, chen2017partial}, when there are multiple modes or the support of the target distribution is concave, singular boundary can emerge. In these regions, the Brenier potential $\boldsymbol{u}_{\boldsymbol{h}}$ is continuous but not differentiable, resulting in a discontinuous gradient map, i.e., the transport map. This indicates that if we extend the OT mapping in these areas, we will generate samples that belong to mixed categories, effectively producing adversarial samples.

The original point cloud requires an abundance of points to represent a single data instance, resulting in high dimensionality~($N\times3$) that complicates the detection of data singular boundaries. To address this, we first embed the point set into a latent representation $\boldsymbol{y}=\mathbf{E}_\phi(\boldsymbol{P})$ on manifold with encoder $\mathbf{E}_\phi$. Next, the core challenge we aim to solve is identifying the discontinuity regions in the OT mapping, which correspond to the singular boundaries.

%Given the OT mapping $T(\cdot)$ solved by SDOT~\cite{an2019ae}, 
Given the OT mapping $T(\cdot)$ solved by  Algorithm~\ref{alg:OTMap}, we can tessellate the data manifold represented by features $\boldsymbol{Y}=\left\{\boldsymbol{y}_i\right\}_{i=1}^{N}$ into $N$ polygons~(illustrated in \cref{fig:illustration}). From a local perspective, each hyperplane $\pi_{\boldsymbol{h},i}$ has boundaries with its neighboring hyperplanes, particularly at the intersections of the two hyperplanes. Some pairs of polygons fall into different categories or exhibit significant normal inconsistencies, indicating that their boundary is singular. Specifically, given $\boldsymbol{y}_i$ from the target domain, we can detect the singular boundaries between it and its neighbors by checking the angles $\theta_{i_k}$ between hyperplane $\pi_{i}$ and $\pi_{i_k}$ with 
\begin{equation}\label{eq:theta}
\theta_{i_k}=\frac{\langle\boldsymbol{y}_{i},\boldsymbol{y}_{i_k}\rangle}{||\boldsymbol{y}_{i}||\cdot||\boldsymbol{y}_{i_k}||},\  k=1,2,\cdots,K. 
\end{equation}
Here, $i_k$ is the index corresponding to the $k$-th neighbor $\boldsymbol{y}_{i_k}$ of $\boldsymbol{y}_i$ which is determined by hyperplane set $\{\pi_{\boldsymbol{h},i}\}$ with 
\begin{equation}
\label{eq:neighbor}
\begin{split}
&\pi_{\boldsymbol{h},i_k}(\boldsymbol{x})\leq\pi_{\boldsymbol{h},i_{k-1}}(\boldsymbol{x})\leq\cdots\leq\pi_{\boldsymbol{h},i_0}(\boldsymbol{x})\leq\pi_{\boldsymbol{h},i}(\boldsymbol{x})\\
&\pi_{\boldsymbol{h},i_k}(\boldsymbol{x})\geq\pi_{\boldsymbol{h},i_{k+1}}(\boldsymbol{x})\geq\cdots\geq\pi_{\boldsymbol{h},i_{N-K-1}}(\boldsymbol{x}).\\
\end{split}
\end{equation}
That is to say, $i_k$ is the index corresponding to the $(k+1)$-th largest hyperplane in $\{\pi_{\boldsymbol{h},i}(\boldsymbol{x})\}^N_{i=1}$ under a random $\boldsymbol{x}$ from $W_{i}$. If there is any angle $\theta_{i_k}$ larger than the given threshold $\tau$, we say $\boldsymbol{x}$ belongs to the singular set, and there is a local singular boundary between $\boldsymbol{y}_{i}$ and  $\boldsymbol{y}_{i_k}$ (solid and dashed red lines in \cref{fig:illustration}).

\begin{table*}[htbp!]
\centering
\renewcommand{\tabcolsep}{0.98mm}
\caption{Comparison results of ASR (\%) for different attack methods with the PointNet++ as the surrogate model to other unknown models.}
\vspace{-3mm}
\scalebox{0.8}{
\begin{tabular}{@{}c|cccc|c|cccc|c@{}}
\toprule
& \multicolumn{5}{c|}{ModelNet40}    & \multicolumn{5}{c}{ShapeNet Part}  \\ 
\midrule
\multirow{2}{*}{Method}  & \multicolumn{4}{c|}{ASR~(\%)$\uparrow$ /\ CD$\downarrow$}  & \multirow{2}{*}{AGT(s)$\downarrow$}  &  \multicolumn{4}{c|}{ASR~(\%)$\uparrow$\ /\ CD$\downarrow$}  & \multirow{2}{*}{AGT(s)$\downarrow$} \\ 
& PointNet  &PointConv &  DGCNN  & PCT &     & PointNet &PointConv & DGCNN & PCT &  \\ 
\midrule
AdvPC~\cite{hamdi2020advpc} & 13.0/0.0005& 30.0/0.0014 & 23.3/0.0011 &  15.8/0.0011 &6.2  & 5.0/0.0024  & 22.5/0.0062&5.7/0.0038 & 6.0/0.0039&  15.8\\
AOF~\cite{liu2022boosting}& 13.7/0.0013& 39.7/0.0035 & 28.1/0.0029 & 18.6/0.0032& 12.5 & 14.6/0.0048& 33.4/0.0063 & 20.1/0.0058& 18.4/0.0058&14.4\\
SI-ADV~\cite{huang2022siadv}  &54.5/0.0022& 69.5/0.0024 &67.3/0.0022 &\textbf{91.3}/0.0026 & 8.9  &19.1/0.0023 &\textbf{77.2}/0.0040 & 18.9/0.0025 &26.9/0.0033&11.4\\
SS-attack~\cite{zhang2024improving} & 15.7/0.0021&44.4/0.0039 &32.0/0.0034 &23.5/0.0038 & 51.5  & 13.0/0.0055  & 43.4/0.0081 & 16.4/0.0065& 22.1/0.0066&43.1\\
HiT-ADV~\cite{lou2024hide}&50.2/0.0330 & 15.0/0.0112&22.3/0.0301 & 9.2/0.0063 &7.5 & 32.4/0.1545 & 7.0/0.0680& 28.7/0.1626& 17.2/0.1890& 25.9\\
NoPain-PF~(ours) &\underline{97.7}/0.0023 &  \underline{72.2}/0.0032&  \underline{88.6}/0.0028& 81.7/0.0029 & \underline{0.028}&\underline{65.2}/0.0022 & 62.5/ 0.0032 & \underline{61.8}/0.0025 & \textbf{60.0}/ 0.0024&\textbf{0.019} \\
NoPain-PD~(ours) & \textbf{100}/0.0022 & \textbf{82.8}/ 0.0024&\textbf{88.7}/ 0.0025 & \underline{85.7}/ 0.0027 & \textbf{0.026} & \textbf{71.1}/ 0.0021 & \underline{63.3}/ 0.0030 & \textbf{75.0}/ 0.0029 & \underline{53.3}/ 0.0046 &\underline{0.032} \\ \bottomrule
\end{tabular}}
\label{tab:transferability}
\end{table*}

\begin{table*}[htbp!]
%\caption{Comparison results of ASR (\%) for different attack methods on PointNet and DGCNN with applied defense methods.}
\caption{Comparison results of ASR (\%) for different attack methods to defense strategies of SRS, SOR, DUP-Net, and IF-Defense.}
\vspace{-3mm}
\centering
\scalebox{0.8}{
\begin{tabular}{@{}c|cccc|cccc@{}}
\toprule
& \multicolumn{4}{c|}{ASR~(\%)$\uparrow$ /\ CD$\downarrow$ ~~~on~ PointNet}    & \multicolumn{4}{c}{ASR~(\%)$\uparrow$ /\ CD$\downarrow$ ~~~on~ DGCNN}  \\ \midrule
Method  & SRS & SOR & DUP-Net &  IF-Defense & SRS & SOR & DUP-Net &  IF-Defense \\ \midrule
AdvPC~\cite{hamdi2020advpc} & 89.5/0.0005 &  34.5/0.0003 & 18.5/0.0003 & 19.3/0.00394  & 63.5/0.0013 & 64.50.0015& 67.5/0.0013&20.8/0.0040 \\
AOF~\cite{liu2022boosting} & 94.0/0.0021
& 88.5/0.0021& 70.5/0.0022&63.7/0.0056  & 52.0/0.0038 & 68.0/0.0026& 70.0/0.0025& 34.1/0.0061\\
SI-ADV~\cite{huang2022siadv} & 86.5/0.0027 & 32.5/0.0029 & 34.5/0.0030& 42.7/0.0040  & \underline{87.5}/0.0034& 68.0/0.0033/& \underline{82.6}/0.0035& 48.4/0.0049\\
SS-attack~\cite{zhang2024improving}  &94.5/0.0023&{89.5}/0.0023 &72.5/0.0025 &  66.1/0.0056  & 71.0/0.0031 &63.5/0.0027 &73.5/0.0023 &41.2/0.0056\\
HiT-ADV~\cite{lou2024hide}& 90.5/0.0736& 86.0/0.0805&  \underline{84.5}/0.0896& 16.2/0.0546 &  55.0/0.1212& 67.5/0.1164& \textbf{87.5}/0.1180&17.2/0.0544 \\
NoPain-PF~(ours) & \underline{97.6}/0.0029 & \underline{90.5}/0.0032 & 83.9/0.0027 & \underline{66.3}/0.0035  &{86.4}/0.0027 & \underline{78.9}/0.0030& 69.7/0.0028 &\underline{50.8}/0.0039   \\
NoPain-PD~(ours) & \textbf{98.4}/0.0021& \textbf{90.7}/0.0024& \textbf{85.0}/0.0028& \textbf{70.0}/0.0033& \textbf{87.9}/0.0026& \textbf{82.8}/0.0028& 74.2/0.0029&\textbf{52.4}/0.0038  \\ \bottomrule
\end{tabular}
}
\label{tab:defend}
\end{table*}

\subsection{Attack with OT Singular Boundary} \label{sec:sampling}
While we can detect singular boundaries, explicitly and accurately calculating them in discrete situations is often intractable or even impossible. Therefore, we extend the semi-discrete OT mapping to obtain the adversarial feature $\boldsymbol{\widehat{y}}$ through the following equation:
%After obtaining the local singular boundary, we can generate adversarial features $\boldsymbol{\widehat{y}}$ by  
\begin{equation}\label{eq:interplatation}
\boldsymbol{\widehat{y}}=\tilde{T}(\boldsymbol{x})=\lambda_iT(\boldsymbol{c}_{i})+\lambda_{i_k}T(\boldsymbol{c}_{i_k})=\lambda_i\boldsymbol{y}_{i}+\lambda_{i_k}\boldsymbol{y}_{i_k}.
\end{equation}
Where the $\mu$-mass center $\boldsymbol{c}_{j}$ is approximated by the mean value of all the Monte-Carlo samples inside $W_{j}$, $\lambda_j=d^{-1}(\boldsymbol{x},\boldsymbol{c}_{j})/(d^{-1}(\boldsymbol{x},\boldsymbol{c}_{i})+d^{-1}(\boldsymbol{x},\boldsymbol{c}_{i_k}))$, $j=i,i_k$. $d^{-1}(\boldsymbol{x},\boldsymbol{c}_{j})$ is the reciprocal of the distance between $\boldsymbol{x}$ and $\boldsymbol{c}_{j}$. $\boldsymbol{x}$ is a random $\boldsymbol{x}$ from $W_{i}$, $\tilde{T}(\cdot)$ is a smoothed extension of the semi-discrete OT mapping $T(\cdot)$, which smooths in regions where latent codes are dense.

Next, we leverage the pre-trained decoder $\mathbf{D}_\varphi$ to generate adversarial samples, denoted as $\boldsymbol{\widehat{P}}=\mathbf{D}_{\varphi}(\boldsymbol{\widehat{y}})$.
The complete attack process is outlined in Algorithm~\ref{alg:ot-attack}.

Thanks to the data manifold decomposition and singular boundary computation, we can efficiently generate adversarial samples using \cref{eq:interplatation} without iterative optimization. Furthermore, our method does not rely on any information from classification models; instead, it directly targets the intrinsic singular boundaries of the data. The adversarial samples generated by sampling within the boundary region exhibit certain unnatural characteristics, which hinder the classification model trained on the original dataset from accurately recognizing them, thereby resulting in cross-network transferability. Additionally, the OT mapping $\tilde{T}(\cdot)$ in \cref{eq:interplatation} is defined by an explicit function, offering geometric intuitiveness and enhancing interpretability.

%%%%%%%%%%%%%%%%%%%%%%%%%%%%%%%%%%%%%%%%%%%%%%%%%%%%%%%%%%%%%%%%%%%%%%%%%%%%%%%%%%%
\section{Experiments}

\subsection{Setup}
\textbf{Dataset. } Following the previous state-of-the-art point cloud attack algorithm \cite{tao20233dhacker,lou2024hide,zhang2024improving}, the experiments in this paper are performed on the ModelNet40~\cite{wu20153dshapenets} and ShapeNetPart~\cite{chang2015shapenet}. % as well as the real-world dataset KITTI~\cite{geiger2012we} and NuScenes~\cite{caesar2020nuscenes}. 
ModelNet40 consists of 12,311 CAD models from 40 categories, of which 9,843 models are intended for training and the other 2,468 for testing. ShapeNetPart consists of 16,881 shapes from 16 categories, split into 12,137 for training and 2,874 for testing. 

\textbf{Baseline attack methods.}
We compare our method with four state-of-the-art attack techniques, including three black-box methods: AdvPC~\cite{hamdi2020advpc}, AOF~\cite{liu2022boosting}, SI-Adv~\cite{huang2022siadv}, and SS-attack~\cite{zhang2024improving}, as well as the white-box method HiT-ADV~\cite{lou2024hide}. We conduct tests using the default settings, official implementations, and pre-training models for all baseline methods for a fair comparison.
Specifically, AdvPC~\cite{hamdi2020advpc} uses an autoencoder to improve transferability. AOF~\cite{liu2022boosting} attacks the more general features of point clouds, thereby improving the transferability of 3D adversarial samples. SI-Adv~\cite{huang2022siadv} introduces perturbations to create shape-invariant point clouds by tangent plane projection. %3DHacker~\cite{tao20233dhacker} is a boundary-based black-box method. PF-Attack~\cite{he2023generating} and ANF~\cite{chen2024anf} optimize perturbations through adversarial noise factorization near decision boundaries. 
SS-attack~\cite{zhang2024improving} applies random scaling or shearing to the input point cloud to prevent overfitting the white-box model. HiT-ADV~\cite{lou2024hide} is a shape-based attack method, that conducts a two-stage search for attack regions based on saliency and imperceptibility scores,
and then adds deformation perturbations in each region using Gaussian kernel functions.

\textbf{Baseline classification models.}
For a fair comparison with baselines, we adopt the same classifiers as AOF and SS-attack, including PointNet~\cite{qi2017pointnet}, PointNet++~\cite{qi2017pointnet++}, DGCNN~\cite{wang2019dynamic}, PointConv~\cite{wu2019pointconv}, and PCT~\cite{guo2021pct}. For the ModelNet40 dataset, we used the pre-trained classification models provided by SS-attack directly for metric evaluation. In contrast, since no pre-trained models were available for the ShapeNetPart dataset, we retrained the classifiers ourselves, achieving final accuracies exceeding 95\% across all models.

\begin{figure*}[htb!]
\centering
\includegraphics[width=0.99\textwidth]{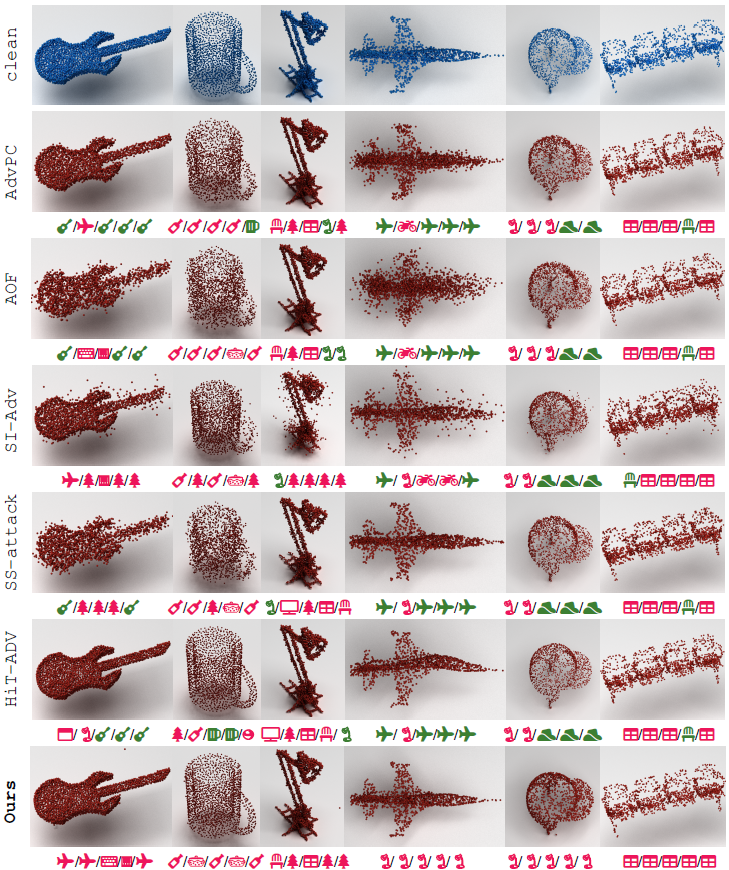}
\caption{Visualizations of adversarial samples on data from ModelNet40~(left three columns) and ShapeNetPart~(right three columns). The icons below point clouds indicate their category prediction by PointNet, PointNet++, PointConv, DGCNN and PCT, where \textcolor{OrangeRed}{red} and \textcolor{OliveGreen}{green} indicate successful and failed attacks.}
\label{fig:qualitative}
\end{figure*}

\textbf{Evaluation metrics} 
%To quantitatively evaluate the proposed method NoPain, we used attack success rate (ASR) to measure attack effectiveness, and chamfer distance (CD)~\cite{fan2017point} to measure the perturbation strength of attack samples. Effective attacks must be based on the premise of limited perturbations, and the success of attacks caused by excessive perturbations is not desirable. Therefore, in subsequent experiments, we will provide the success rate of attacks and the corresponding cd value. Moreover, we compare the Average Time Cost (ATC) in seconds for generating an adversarial sample using an NVIDIA A40 GPU. 
To quantitatively evaluate our proposed method, NoPain, we used Attack Success Rate (ASR) to assess attack effectiveness and Chamfer Distance (CD)~\cite{fan2017point} to measure the perturbation strength of adversarial samples. Effective attacks require limited perturbations, as the success achieved with excessive perturbations is undesirable. Therefore, in the following experiments, we report both the ASR and corresponding CD values. To assess transferability, we tested the adversarial samples on various target classification models; high ASR scores across models indicate strong transferability. Additionally, we compare the Average Time Cost (ATC) in seconds for generating each adversarial sample on an NVIDIA A40 GPU.

\iffalse
The definitions of ASR and TSR are as follows:
\begin{align*}
&ASR = \frac{n_s}{n_{t}}\\
&TSR = \frac{n_b}{n_s},
\end{align*}
where $n_{t}$ is the total number of generated adversarial samples. $n_{s}$ is the number of successful attack samples on the victim model. $n_b$ represents the number of successful attack samples on both the victim and test model.
\fi

\textbf{Implementation details.}
Our attack framework NoPain is adaptable to various autoencoder architectures. We demonstrate its effectiveness using two widely recognized point cloud autoencoders: PointFlow~\cite{yang2019pointflow} and Point-Diffusion~\cite{luo2021diffusion}, denoted as \textbf{NoPain-PF} and \textbf{NoPain-PD} respectively. PointFlow leverages continuous normalizing flows to transform simple distributions into complex point cloud distributions through a series of invertible transformations, enabling precise point cloud generation and meaningful latent space interpolation. Point-Diffusion, on the other hand, is a diffusion-based approach that excels in generating diverse, high-fidelity point clouds. Both models provide pre-trained encoders and decoders for the target dataset, making them particularly suitable for our framework.

In Algorithm~\ref{alg:OTMap}, we set $M=10N$, initial learning rate $lr=10^{-2}$, threshold $\eta=2\times10^{-3}$, $s=50$.
In Algorithm~\ref{alg:ot-attack}, we set $K=11$ and $\tau=1.6$ on ModelNet40, and set $K=11$ and $\tau=0.9$ on ShapeNetPart.

\subsection{Quantitative Results} 
\noindent\textbf{Transferability.} We report the Attack Success Rate (ASR) against four target models and the Chamfer Distance (CD) between successfully attacked samples and the original samples in \cref{tab:transferability}. For all baseline methods, adversarial samples were generated using PointNet++ as the surrogate model. The results indicate that supervised black-box AdvPC, AOF, SI-Adv, and SS-attack, rely on model-specific loss, resulting in lower ASR on target models. In contrast, the white-box HiT-ADV shows even lower transferability due to its strong dependence on model-specific information.

Compared to these baselines, our methods, NoPain-PF and NoPain-PD, produce adversarial samples with comparable perturbations (CD) and achieve consistently high ASR across all four classification models. Notably, the diffusion-based NoPain-PD attains the highest ASR on most classifiers, indicating that our adversarial samples exhibit strong transferability. Leveraging OT, our approach can detect the singular boundaries of the data manifold, sampling along these boundaries to generate mode-mixed adversarial samples and facilitate transferable end-to-end attacks.

Furthermore, as measured by the AGT metric, our method is an efficient end-to-end approach that only requires a single-step OT mapping to generate adversarial samples, significantly reducing computational costs.

\textbf{Attack against Defense.} To evaluate the robustness of our proposed NoPain under various 3D adversarial defense algorithms, we conducted tests on classification models with four different defense methods, i.e. SRS, SOR, DUP-Net~\cite{zhou2019dup} and  %AT~\cite{liu2019extending}.
IF-Defense~\cite{liu2019extending}. For IF-Defense, we adopt the ConveNet~\cite{peng2020convolutional} model for defense. The defense algorithms in this paper are all implemented using the open-source code provided by IF-Defense.

We generate adversarial examples and calculate the ASR on victim models with defenses. For evaluation, PointNet and DGCNN are selected as the victim models, and the experimental results are reported in Tab.~\ref{tab:defend}. Our method demonstrates robust attack performance across all models while maintaining comparable CD scores. This robustness stems from our approach, which targets the intrinsic characteristics of the data manifold, i.e. the singular boundaries, to produce adversarial examples that are not commonly seen in the original dataset, making them challenging for overfitted classifiers to accurately identify.

\subsection{Qualitative Results} 
The adversarial point clouds generated by different methods are shown in \cref{fig:qualitative}. Here, all baseline methods AdvPC, AOF, SI-Adv, SS-attack, and HiT-ADV adopt Pointnet++ as the surrogate model. These results reveal that baseline methods face challenges in achieving effective network-transferable attacks. In contrast, our NoPain exhibits robust transferability across classifiers. Our model successfully attacks five classifiers simultaneously, whereas other baseline methods tend to achieve high success rates only on surrogate models or those with similar structures. When there is a significant difference between the test and surrogate classifiers, e.g. PointNet++ and PCT, baseline methods often struggle to induce misclassification.
Especially for the more complex ShapeNetPart, the transferability of baselines is even worse, such as on the airplane in the fourth column.

\subsection{Ablation studies} 
\begin{figure}[htb!]
\vspace{-4mm}
\centering
\includegraphics[width=0.495\linewidth]{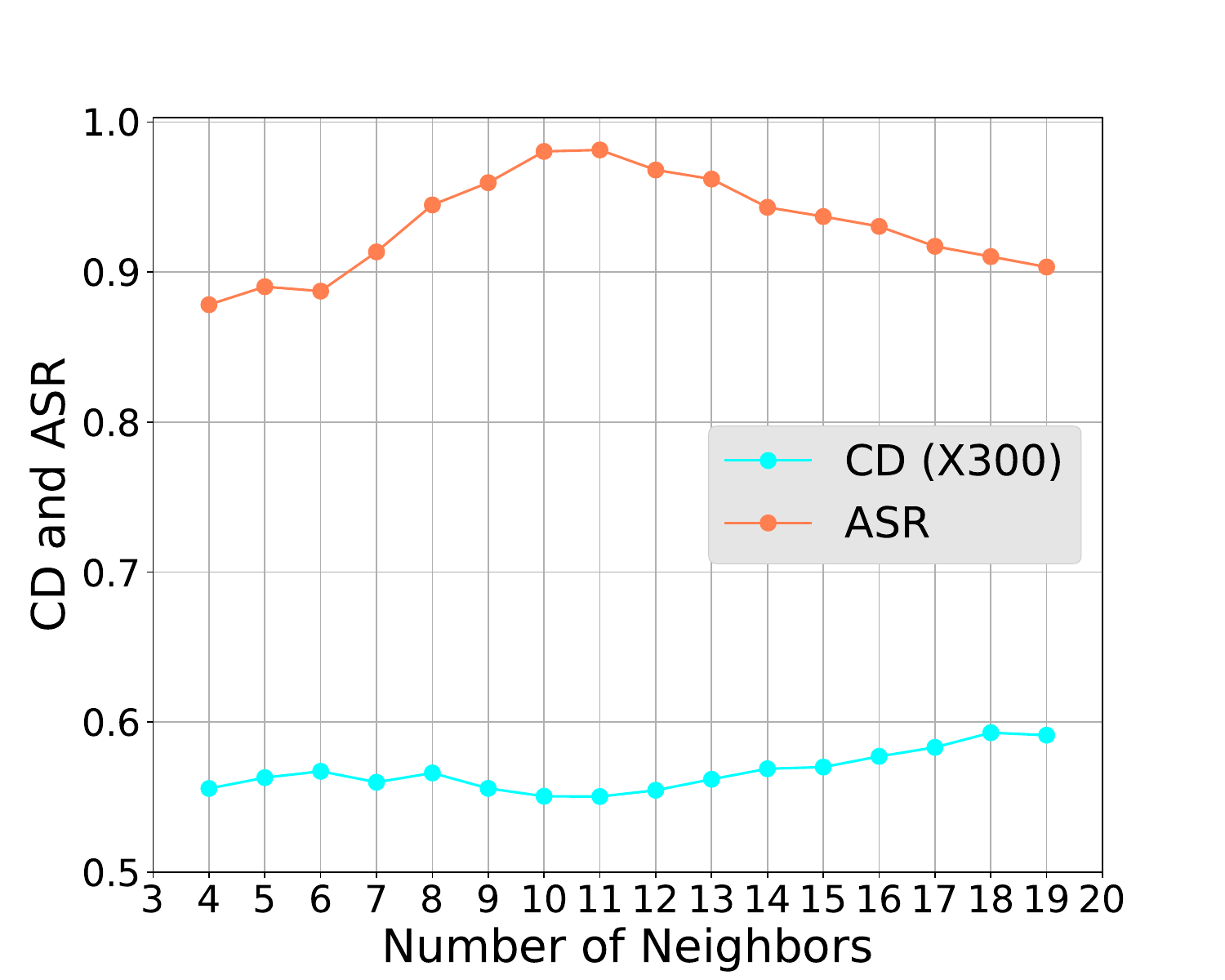}
\includegraphics[width=0.495\linewidth]{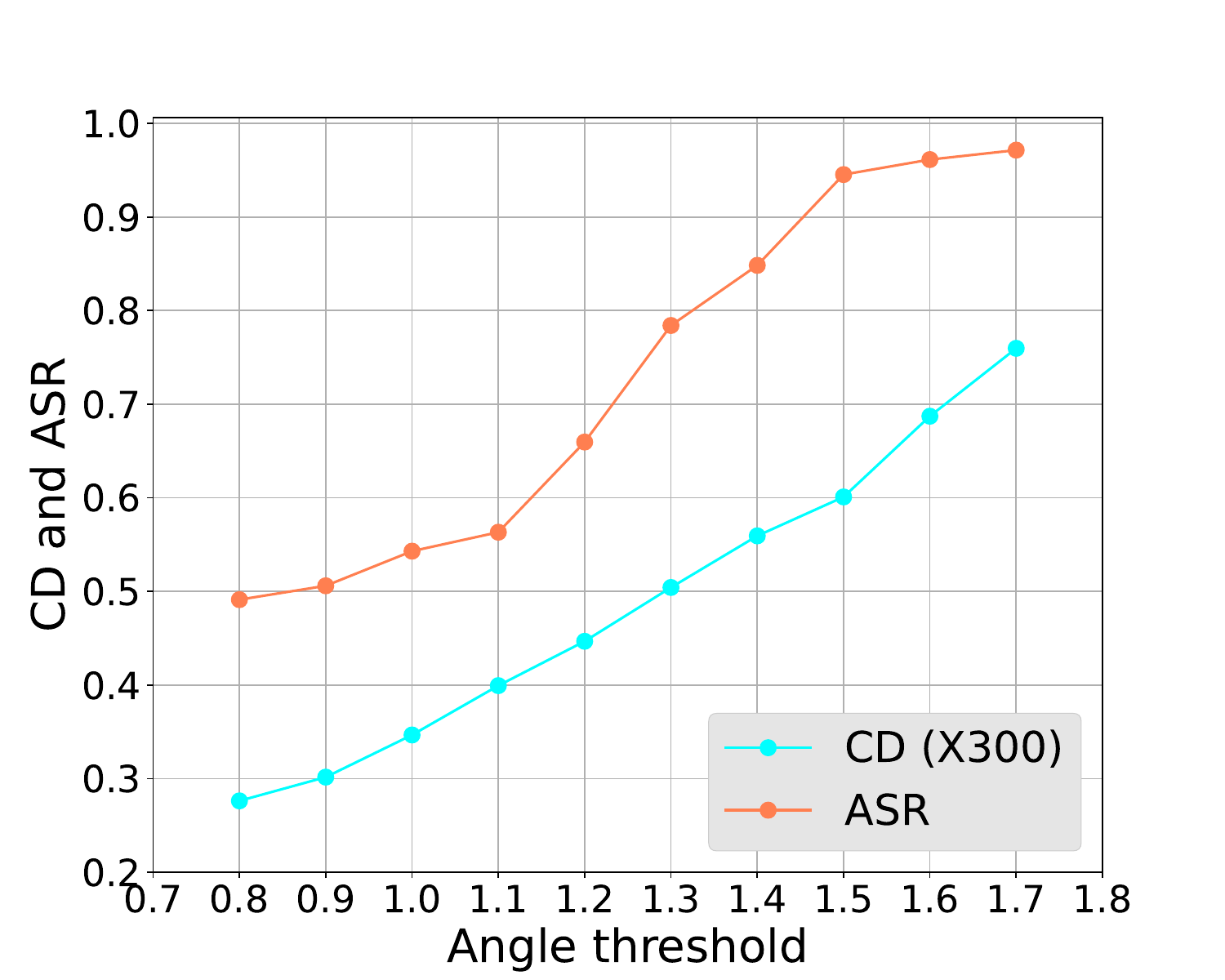}
%\vspace{-2mm}
\caption{Effects of the number of neighbors 
$K$ and angle threshold $\tau$ to ASR and CD on ModelNet40. To present these two metrics in a single graph, we scaled the CD values by a factor of 300.}
\label{fig:ablation}
\end{figure}

To validate the effectiveness of specific hyperparameter settings in our method, we conducted ablation studies on the number of neighbors $K$ and threshold $\tau$ in Algorithm~\ref{alg:ot-attack}, using PointNet as the victim model. The experimental results are presented in Fig.~\ref{fig:ablation}. The graph on the left shows that $K$ reaches its optimum at 10 and 11, where the attack success rate (ASR) is highest and the Chamfer distance (CD) is lowest. The graph on the right indicates that as $\tau$ increases, both ASR and CD rise simultaneously. To constrain the perturbations of adversarial samples, we set $\tau$ to 1.6 for the experiment, achieving an ASR of 97\%.
\vspace{-2mm}
%%%%%%%%%%%%%%%%%%%%%%%%%%%%%%%%%%%%%%%%%%%%%%%%%%%%%%%%%%%%%%%%%%%%%%%%%%%%%%%%%%%
\section{Conclusion}
In this paper, we introduced NoPain, a novel and interpretable adversarial attack framework that leverages OT to identify singular boundaries within data. Unlike traditional approaches, NoPain generates transferable adversarial examples without requiring iterative updates or guidance from surrogate models. By solving the OT mapping from noise to feature space, our method determined singular boundaries on the target data manifold and shifted point cloud features toward these boundaries to execute the attack. This strategy not only enhanced the interpretability of the approach but also eliminated reliance on classifiers. Experimental results demonstrated the effectiveness of our no-box attack algorithm, with NoPain producing adversarial samples that offer superior transferability and efficiency over existing methods, as confirmed by extensive comparative experiments.

\section*{Acknowledgment}
This research was supported by the National Key R$\&$D Program of China under Grant No. 2021YFA1003003, the Natural Science Foundation of China under Grant No. 62306059 and No. T2225012.  
%\newpage
{\small
\bibliographystyle{ieee_fullname}
\bibliography{ref}

\begin{thebibliography}{66}
\providecommand{\natexlab}[1]{#1}
\providecommand{\url}[1]{\texttt{#1}}
\expandafter\ifx\csname urlstyle\endcsname\relax
  \providecommand{\doi}[1]{doi: #1}\else
  \providecommand{\doi}{doi: \begingroup \urlstyle{rm}\Url}\fi

\bibitem[An et~al.(2019)An, Guo, Lei, Luo, Yau, and Gu]{an2019ae}
Dongsheng An, Yang Guo, Na Lei, Zhongxuan Luo, Shing-Tung Yau, and Xianfeng Gu.
\newblock Ae-ot: a new generative model based on extended semi-discrete optimal
  transport.
\newblock \emph{ICLR 2020}, 2019.

\bibitem[Athalye et~al.(2018)Athalye, Carlini, and
  Wagner]{athalye2018obfuscated}
Anish Athalye, Nicholas Carlini, and David Wagner.
\newblock Obfuscated gradients give a false sense of security: Circumventing
  defenses to adversarial examples.
\newblock In \emph{International conference on machine learning}, pages
  274--283. PMLR, 2018.

\bibitem[Brendel et~al.(2018)Brendel, Rauber, and Bethge]{brendel2018decision}
Wieland Brendel, Jonas Rauber, and Matthias Bethge.
\newblock Decision-based adversarial attacks: Reliable attacks against
  black-box machine learning models.
\newblock In \emph{International Conference on Learning Representations}, 2018.

\bibitem[Brenier(1991)]{Brenier1991}
Yann Brenier.
\newblock Polar factorization and monotone rearrangement of vector-valued
  functions.
\newblock \emph{Communications on pure and applied mathematics}, 44\penalty0
  (4):\penalty0 375--417, 1991.

\bibitem[Brunner et~al.(2019)Brunner, Diehl, Le, and
  Knoll]{brunner2019guessing}
Thomas Brunner, Frederik Diehl, Michael~Truong Le, and Alois Knoll.
\newblock Guessing smart: Biased sampling for efficient black-box adversarial
  attacks.
\newblock In \emph{Proceedings of the IEEE/CVF International Conference on
  Computer Vision}, pages 4958--4966, 2019.

\bibitem[Cai et~al.(2024)Cai, Tao, Liu, Zhou, Qu, Dong, Tang, and
  Sun]{cai2024frequency}
Xiaowen Cai, Yunbo Tao, Daizong Liu, Pan Zhou, Xiaoye Qu, Jianfeng Dong, Keke
  Tang, and Lichao Sun.
\newblock Frequency-aware gan for imperceptible transfer attack on 3d point
  clouds.
\newblock In \emph{Proceedings of the 32nd ACM International Conference on
  Multimedia}, pages 6162--6171, 2024.

\bibitem[Carlini and Wagner(2017)]{carlini2017cw}
Nicholas Carlini and David Wagner.
\newblock Towards evaluating the robustness of neural networks.
\newblock In \emph{2017 ieee symposium on security and privacy (sp)}, pages
  39--57. Ieee, 2017.

\bibitem[Chang et~al.(2015)Chang, Funkhouser, Guibas, Hanrahan, Huang, Li,
  Savarese, Savva, Song, Su, et~al.]{chang2015shapenet}
Angel~X Chang, Thomas Funkhouser, Leonidas Guibas, Pat Hanrahan, Qixing Huang,
  Zimo Li, Silvio Savarese, Manolis Savva, Shuran Song, Hao Su, et~al.
\newblock Shapenet: An information-rich 3d model repository.
\newblock \emph{arXiv preprint arXiv:1512.03012}, 2015.

\bibitem[Chen et~al.(2024)Chen, Zhao, Yang, Yan, He, Xue, Qian, and
  Su]{chen2024anf}
Hai Chen, Shu Zhao, Xiao Yang, Huanqian Yan, Yuan He, Hui Xue, Fulan Qian, and
  Hang Su.
\newblock Anf: Crafting transferable adversarial point clouds via adversarial
  noise factorization.
\newblock \emph{IEEE Transactions on Big Data}, 2024.

\bibitem[Chen et~al.(2020)Chen, Jordan, and
  Wainwright]{chen2020hopskipjumpattack}
Jianbo Chen, Michael~I Jordan, and Martin~J Wainwright.
\newblock Hopskipjumpattack: A query-efficient decision-based attack.
\newblock In \emph{2020 ieee symposium on security and privacy (sp)}, pages
  1277--1294. IEEE, 2020.

\bibitem[Chen and Figalli(2017)]{chen2017partial}
Shibing Chen and Alessio Figalli.
\newblock Partial w2, p regularity for optimal transport maps.
\newblock \emph{Journal of Functional Analysis}, 272\penalty0 (11):\penalty0
  4588--4605, 2017.

\bibitem[Cheng et~al.(2021)Cheng, Sang, Zhou, and Wang]{cheng2021universal}
Riran Cheng, Nan Sang, Yinyuan Zhou, and Xupeng Wang.
\newblock Universal adversarial attack against 3d object tracking.
\newblock In \emph{2021 IEEE 23rd Int Conf on High Performance Computing \&
  Communications; 7th Int Conf on Data Science \& Systems; 19th Int Conf on
  Smart City; 7th Int Conf on Dependability in Sensor, Cloud \& Big Data
  Systems \& Application (HPCC/DSS/SmartCity/DependSys)}, pages 34--40. IEEE,
  2021.

\bibitem[Dong et~al.(2022)Dong, Zhu, Gao, et~al.]{dong2022isometric}
Yinpeng Dong, Jun Zhu, Xiao-Shan Gao, et~al.
\newblock Isometric 3d adversarial examples in the physical world.
\newblock \emph{Advances in Neural Information Processing Systems},
  35:\penalty0 19716--19731, 2022.

\bibitem[Duboudin et~al.(2022)Duboudin, Dellandr{\'e}a, Abgrall, H{\'e}naff,
  and Chen]{duboudin2022look}
Thomas Duboudin, Emmanuel Dellandr{\'e}a, Corentin Abgrall, Gilles H{\'e}naff,
  and Liming Chen.
\newblock Look beyond bias with entropic adversarial data augmentation.
\newblock In \emph{2022 26th International Conference on Pattern Recognition
  (ICPR)}, pages 2142--2148. IEEE, 2022.

\bibitem[Fan et~al.(2017)Fan, Su, and Guibas]{fan2017point}
Haoqiang Fan, Hao Su, and Leonidas~J Guibas.
\newblock A point set generation network for 3d object reconstruction from a
  single image.
\newblock In \emph{Proceedings of the IEEE conference on computer vision and
  pattern recognition}, pages 605--613, 2017.

\bibitem[Figalli(2010)]{figalli2010regularity}
Alessio Figalli.
\newblock Regularity properties of optimal maps between nonconvex domains in
  the plane.
\newblock \emph{Communications in Partial Differential Equations}, 35\penalty0
  (3):\penalty0 465--479, 2010.

\bibitem[Goodfellow et~al.(2014)Goodfellow, Shlens, and
  Szegedy]{goodfellow2014FGSM}
Ian~J Goodfellow, Jonathon Shlens, and Christian Szegedy.
\newblock Explaining and harnessing adversarial examples.
\newblock \emph{arXiv preprint arXiv:1412.6572}, 2014.

\bibitem[Guo et~al.(2021)Guo, Cai, Liu, Mu, Martin, and Hu]{guo2021pct}
Meng-Hao Guo, Jun-Xiong Cai, Zheng-Ning Liu, Tai-Jiang Mu, Ralph~R Martin, and
  Shi-Min Hu.
\newblock Pct: Point cloud transformer.
\newblock \emph{Computational Visual Media}, 7:\penalty0 187--199, 2021.

\bibitem[Hamdi et~al.(2020)Hamdi, Rojas, Thabet, and Ghanem]{hamdi2020advpc}
Abdullah Hamdi, Sara Rojas, Ali Thabet, and Bernard Ghanem.
\newblock Advpc: Transferable adversarial perturbations on 3d point clouds.
\newblock In \emph{Computer Vision--ECCV 2020: 16th European Conference,
  Glasgow, UK, August 23--28, 2020, Proceedings, Part XII 16}, pages 241--257.
  Springer, 2020.

\bibitem[Han et~al.(2023)Han, Jia, Bai, Gu, Liu, and Cao]{han2023ot}
Dongchen Han, Xiaojun Jia, Yang Bai, Jindong Gu, Yang Liu, and Xiaochun Cao.
\newblock Ot-attack: Enhancing adversarial transferability of vision-language
  models via optimal transport optimization.
\newblock \emph{arXiv preprint arXiv:2312.04403}, 2023.

\bibitem[He et~al.(2023)He, Liu, Li, Liang, Li, Jia, and Cao]{he2023generating}
Bangyan He, Jian Liu, Yiming Li, Siyuan Liang, Jingzhi Li, Xiaojun Jia, and
  Xiaochun Cao.
\newblock Generating transferable 3d adversarial point cloud via random
  perturbation factorization.
\newblock In \emph{Proceedings of the AAAI Conference on Artificial
  Intelligence}, pages 764--772, 2023.

\bibitem[Huang et~al.(2022)Huang, Dong, Chen, Zhou, Zhang, and
  Yu]{huang2022siadv}
Qidong Huang, Xiaoyi Dong, Dongdong Chen, Hang Zhou, Weiming Zhang, and Nenghai
  Yu.
\newblock Shape-invariant 3d adversarial point clouds.
\newblock In \emph{Proceedings of the IEEE/CVF Conference on Computer Vision
  and Pattern Recognition}, pages 15335--15344, 2022.

\bibitem[Kim et~al.(2021)Kim, Hua, Nguyen, and Yeung]{kim2021minimal}
Jaeyeon Kim, Binh-Son Hua, Thanh Nguyen, and Sai-Kit Yeung.
\newblock Minimal adversarial examples for deep learning on 3d point clouds.
\newblock In \emph{Proceedings of the IEEE/CVF International Conference on
  Computer Vision}, pages 7797--7806, 2021.

\bibitem[Kingma and Ba(2014)]{adam2014}
Diederik~P Kingma and Jimmy Ba.
\newblock Adam: A method for stochastic optimization.
\newblock \emph{arXiv preprint arXiv:1412.6980}, 2014.

\bibitem[Labarbarie et~al.(2024)Labarbarie, Chan-Hon-Tong, Herbin, and
  Leyli-Abadi]{labarbarie2024optimal}
Pol Labarbarie, Adrien Chan-Hon-Tong, St{\'e}phane Herbin, and Milad
  Leyli-Abadi.
\newblock Optimal transport based adversarial patch to leverage large scale
  attack transferability.
\newblock In \emph{The International Conference on Learning Representations
  (ICLR 2024)}, 2024.

\bibitem[Lei et~al.(2020)Lei, An, Guo, Su, Liu, Luo, Yau, and
  Gu]{lei2020geometric}
Na Lei, Dongsheng An, Yang Guo, Kehua Su, Shixia Liu, Zhongxuan Luo, Shing-Tung
  Yau, and Xianfeng Gu.
\newblock A geometric understanding of deep learning.
\newblock \emph{Engineering}, 6\penalty0 (3):\penalty0 361--374, 2020.

\bibitem[Li et~al.(2020{\natexlab{a}})Li, Xu, Zhang, Yang, and Li]{li2020qeba}
Huichen Li, Xiaojun Xu, Xiaolu Zhang, Shuang Yang, and Bo Li.
\newblock Qeba: Query-efficient boundary-based blackbox attack.
\newblock In \emph{Proceedings of the IEEE/CVF conference on computer vision
  and pattern recognition}, pages 1221--1230, 2020{\natexlab{a}}.

\bibitem[Li et~al.(2021)Li, Li, Xu, Zhang, Yang, and Li]{li2021nonlinear}
Huichen Li, Linyi Li, Xiaojun Xu, Xiaolu Zhang, Shuang Yang, and Bo Li.
\newblock Nonlinear projection based gradient estimation for query efficient
  blackbox attacks.
\newblock In \emph{International Conference on Artificial Intelligence and
  Statistics}, pages 3142--3150. PMLR, 2021.

\bibitem[Li et~al.(2020{\natexlab{b}})Li, Guo, and Chen]{li2020practical}
Qizhang Li, Yiwen Guo, and Hao Chen.
\newblock Practical no-box adversarial attacks against dnns.
\newblock \emph{Advances in Neural Information Processing Systems},
  33:\penalty0 12849--12860, 2020{\natexlab{b}}.

\bibitem[Li et~al.(2022)Li, Zhang, Yin, and Liu]{li2022decision}
Xiu-Chuan Li, Xu-Yao Zhang, Fei Yin, and Cheng-Lin Liu.
\newblock Decision-based adversarial attack with frequency mixup.
\newblock \emph{IEEE Transactions on Information Forensics and Security},
  17:\penalty0 1038--1052, 2022.

\bibitem[Li et~al.(2024)Li, Li, Jin, Lei, and Luo]{li2023ot}
Zezeng Li, Shenghao Li, Lianbao Jin, Na Lei, and Zhongxuan Luo.
\newblock Ot-net: A reusable neural optimal transport solver.
\newblock \emph{Machine Learning}, pages 1--26, 2024.

\bibitem[Liu et~al.(2022)Liu, Zhang, and Zhu]{liu2022boosting}
Binbin Liu, Jinlai Zhang, and Jihong Zhu.
\newblock Boosting 3d adversarial attacks with attacking on frequency.
\newblock \emph{IEEE Access}, 10:\penalty0 50974--50984, 2022.

\bibitem[Liu and Hu(2022)]{liu2022imperceptible}
Daizong Liu and Wei Hu.
\newblock Imperceptible transfer attack and defense on 3d point cloud
  classification.
\newblock \emph{IEEE transactions on pattern analysis and machine
  intelligence}, 45\penalty0 (4):\penalty0 4727--4746, 2022.

\bibitem[Liu et~al.(2019)Liu, Yu, and Su]{liu2019extending}
Daniel Liu, Ronald Yu, and Hao Su.
\newblock Extending adversarial attacks and defenses to deep 3d point cloud
  classifiers.
\newblock In \emph{2019 IEEE International Conference on Image Processing
  (ICIP)}, pages 2279--2283. IEEE, 2019.

\bibitem[Liu et~al.(2020)Liu, Yu, and Su]{liu2020sink}
Daniel Liu, Ronald Yu, and Hao Su.
\newblock Adversarial shape perturbations on 3d point clouds.
\newblock In \emph{Computer Vision--ECCV 2020 Workshops: Glasgow, UK, August
  23--28, 2020, Proceedings, Part I 16}, pages 88--104. Springer, 2020.

\bibitem[Lou et~al.(2024)Lou, Jia, Gu, Liu, Liang, He, and Cao]{lou2024hide}
Tianrui Lou, Xiaojun Jia, Jindong Gu, Li Liu, Siyuan Liang, Bangyan He, and
  Xiaochun Cao.
\newblock Hide in thicket: Generating imperceptible and rational adversarial
  perturbations on 3d point clouds.
\newblock In \emph{Proceedings of the IEEE/CVF Conference on Computer Vision
  and Pattern Recognition}, pages 24326--24335, 2024.

\bibitem[Lu et~al.(2023)Lu, Wang, Chang, Yang, and Shum]{lu2023hard}
Zhengzhi Lu, He Wang, Ziyi Chang, Guoan Yang, and Hubert~PH Shum.
\newblock Hard no-box adversarial attack on skeleton-based human action
  recognition with skeleton-motion-informed gradient.
\newblock In \emph{Proceedings of the IEEE/CVF International Conference on
  Computer Vision}, pages 4597--4606, 2023.

\bibitem[Luo and Hu(2021)]{luo2021diffusion}
Shitong Luo and Wei Hu.
\newblock Diffusion probabilistic models for 3d point cloud generation.
\newblock In \emph{Proceedings of the IEEE/CVF conference on computer vision
  and pattern recognition}, pages 2837--2845, 2021.

\bibitem[Moosavi-Dezfooli et~al.(2016)Moosavi-Dezfooli, Fawzi, and
  Frossard]{moosavi2016deepfool}
Seyed-Mohsen Moosavi-Dezfooli, Alhussein Fawzi, and Pascal Frossard.
\newblock Deepfool: a simple and accurate method to fool deep neural networks.
\newblock In \emph{Proceedings of the IEEE conference on computer vision and
  pattern recognition}, pages 2574--2582, 2016.

\bibitem[Mou et~al.(2024)Mou, Guo, Zhao, Wang, Zhao, and Wang]{mou2024no}
Ningping Mou, Binqing Guo, Lingchen Zhao, Cong Wang, Yue Zhao, and Qian Wang.
\newblock No-box universal adversarial perturbations against image classifiers
  via artificial textures.
\newblock \emph{IEEE Transactions on Information Forensics and Security}, 2024.

\bibitem[Papernot et~al.(2016)Papernot, McDaniel, Jha, Fredrikson, Celik, and
  Swami]{papernot2016limitations}
Nicolas Papernot, Patrick McDaniel, Somesh Jha, Matt Fredrikson, Z~Berkay
  Celik, and Ananthram Swami.
\newblock The limitations of deep learning in adversarial settings.
\newblock In \emph{2016 IEEE European symposium on security and privacy
  (EuroS\&P)}, pages 372--387. IEEE, 2016.

\bibitem[Peng et~al.(2020)Peng, Niemeyer, Mescheder, Pollefeys, and
  Geiger]{peng2020convolutional}
Songyou Peng, Michael Niemeyer, Lars Mescheder, Marc Pollefeys, and Andreas
  Geiger.
\newblock Convolutional occupancy networks.
\newblock In \emph{Computer Vision--ECCV 2020: 16th European Conference,
  Glasgow, UK, August 23--28, 2020, Proceedings, Part III 16}, pages 523--540.
  Springer, 2020.

\bibitem[Qi et~al.(2017{\natexlab{a}})Qi, Su, Mo, and Guibas]{qi2017pointnet}
Charles~R Qi, Hao Su, Kaichun Mo, and Leonidas~J Guibas.
\newblock Pointnet: Deep learning on point sets for 3d classification and
  segmentation.
\newblock In \emph{Proceedings of the IEEE Conference on Computer Vision and
  Pattern Recognition (CVPR)}, pages 652--660, 2017{\natexlab{a}}.

\bibitem[Qi et~al.(2017{\natexlab{b}})Qi, Yi, Su, and Guibas]{qi2017pointnet++}
Charles~R Qi, Li Yi, Hao Su, and Leonidas~J Guibas.
\newblock Pointnet++: Deep hierarchical feature learning on point sets in a
  metric space.
\newblock In \emph{Advances in Neural Information Processing Systems (NIPS)},
  2017{\natexlab{b}}.

\bibitem[Rampini et~al.(2021)Rampini, Pestarini, Cosmo, Melzi, and
  Rodola]{rampini2021universal}
Arianna Rampini, Franco Pestarini, Luca Cosmo, Simone Melzi, and Emanuele
  Rodola.
\newblock Universal spectral adversarial attacks for deformable shapes.
\newblock In \emph{Proceedings of the IEEE/CVF conference on computer vision
  and pattern recognition}, pages 3216--3226, 2021.

\bibitem[Shi et~al.(2022)Shi, Chen, Xu, Yang, Yu, and Huang]{shi2022shape}
Zhenbo Shi, Zhi Chen, Zhenbo Xu, Wei Yang, Zhidong Yu, and Liusheng Huang.
\newblock Shape prior guided attack: Sparser perturbations on 3d point clouds.
\newblock In \emph{Proceedings of the AAAI Conference on Artificial
  Intelligence}, pages 8277--8285, 2022.

\bibitem[Srinivasan et~al.(2019)Srinivasan, Kuruoglu, M{\"u}ller, Samek, and
  Nakajima]{srinivasan2019black}
Vignesh Srinivasan, Ercan~E Kuruoglu, Klaus-Robert M{\"u}ller, Wojciech Samek,
  and Shinichi Nakajima.
\newblock Black-box decision based adversarial attack with symmetric
  $\alpha$-stable distribution.
\newblock In \emph{2019 27th European Signal Processing Conference (EUSIPCO)},
  pages 1--5. IEEE, 2019.

\bibitem[Sun et~al.(2022)Sun, Zhang, Chaoqun, Wang, Li, Liu, Han, and
  Tian]{sun2022towards}
Chenghao Sun, Yonggang Zhang, Wan Chaoqun, Qizhou Wang, Ya Li, Tongliang Liu,
  Bo Han, and Xinmei Tian.
\newblock Towards lightweight black-box attack against deep neural networks.
\newblock \emph{Advances in Neural Information Processing Systems},
  35:\penalty0 19319--19331, 2022.

\bibitem[Sun et~al.(2021)Sun, Cao, Choy, Yu, Anandkumar, Mao, and
  Xiao]{sun2021adversarially}
Jiachen Sun, Yulong Cao, Christopher~B Choy, Zhiding Yu, Anima Anandkumar,
  Zhuoqing~Morley Mao, and Chaowei Xiao.
\newblock Adversarially robust 3d point cloud recognition using
  self-supervisions.
\newblock \emph{Advances in Neural Information Processing Systems},
  34:\penalty0 15498--15512, 2021.

\bibitem[Szegedy(2013)]{szegedy2013intriguing}
C Szegedy.
\newblock Intriguing properties of neural networks.
\newblock \emph{arXiv preprint arXiv:1312.6199}, 2013.

\bibitem[Tang et~al.(2023)Tang, Wu, Peng, Shi, Song, Gu, Tian, and
  Wang]{tang2023manifold}
Keke Tang, Jianpeng Wu, Weilong Peng, Yawen Shi, Peng Song, Zhaoquan Gu,
  Zhihong Tian, and Wenping Wang.
\newblock Deep manifold attack on point clouds via parameter plane stretching.
\newblock In \emph{Proceedings of the AAAI Conference on Artificial
  Intelligence}, pages 2420--2428, 2023.

\bibitem[Tao et~al.(2023)Tao, Liu, Zhou, Xie, Du, and Hu]{tao20233dhacker}
Yunbo Tao, Daizong Liu, Pan Zhou, Yulai Xie, Wei Du, and Wei Hu.
\newblock 3dhacker: Spectrum-based decision boundary generation for hard-label
  3d point cloud attack.
\newblock In \emph{Proceedings of the IEEE/CVF International Conference on
  Computer Vision}, pages 14340--14350, 2023.

\bibitem[Tsai et~al.(2020)Tsai, Yang, Ho, and Jin]{tsai2020robust}
Tzungyu Tsai, Kaichen Yang, Tsung-Yi Ho, and Yier Jin.
\newblock Robust adversarial objects against deep learning models.
\newblock In \emph{Proceedings of the AAAI Conference on Artificial
  Intelligence}, pages 954--962, 2020.

\bibitem[Wang et~al.(2019)Wang, Sun, Liu, Sarma, Bronstein, and
  Solomon]{wang2019dynamic}
Yue Wang, Yongbin Sun, Ziwei Liu, Sanjay~E Sarma, Michael~M Bronstein, and
  Justin~M Solomon.
\newblock Dynamic graph cnn for learning on point clouds.
\newblock \emph{ACM Transactions on Graphics (tog)}, 38\penalty0 (5):\penalty0
  1--12, 2019.

\bibitem[Wen et~al.(2020)Wen, Lin, Chen, Chen, and Jia]{wen2020geometry}
Yuxin Wen, Jiehong Lin, Ke Chen, CL~Philip Chen, and Kui Jia.
\newblock Geometry-aware generation of adversarial point clouds.
\newblock \emph{IEEE Transactions on Pattern Analysis and Machine
  Intelligence}, 44\penalty0 (6):\penalty0 2984--2999, 2020.

\bibitem[Wu et~al.(2019)Wu, Qi, and Fuxin]{wu2019pointconv}
Wenxuan Wu, Zhongang Qi, and Li Fuxin.
\newblock Pointconv: Deep convolutional networks on 3d point clouds.
\newblock In \emph{Proceedings of the IEEE/CVF Conference on computer vision
  and pattern recognition}, pages 9621--9630, 2019.

\bibitem[Wu et~al.(2015)Wu, Song, Khosla, Yu, Zhang, Tang, and
  Xiao]{wu20153dshapenets}
Zhirong Wu, Shuran Song, Aditya Khosla, Fisher Yu, Linguang Zhang, Xiaoou Tang,
  and Jianxiong Xiao.
\newblock 3d shapenets: A deep representation for volumetric shapes.
\newblock In \emph{Proceedings of the IEEE Conference on Computer Vision and
  Pattern Recognition (CVPR)}, pages 1912--1920, 2015.

\bibitem[Xiang et~al.(2019)Xiang, Qi, and Li]{xiang2019generating}
Chong Xiang, Charles~R Qi, and Bo Li.
\newblock Generating 3d adversarial point clouds.
\newblock In \emph{Proceedings of the IEEE/CVF Conference on Computer Vision
  and Pattern Recognition}, pages 9136--9144, 2019.

\bibitem[Yang et~al.(2019)Yang, Huang, Hao, Liu, Belongie, and
  Hariharan]{yang2019pointflow}
Guandao Yang, Xun Huang, Zekun Hao, Ming-Yu Liu, Serge Belongie, and Bharath
  Hariharan.
\newblock Pointflow: 3d point cloud generation with continuous normalizing
  flows.
\newblock In \emph{Proceedings of the IEEE/CVF international conference on
  computer vision}, pages 4541--4550, 2019.

\bibitem[Zhang et~al.(2024{\natexlab{a}})Zhang, Cheng, He, Huang, Li, Sicre,
  Huang, Hermanns, and Zhang]{zhang2024eidos}
Hanwei Zhang, Luo Cheng, Qisong He, Wei Huang, Renjue Li, Ronan Sicre, Xiaowei
  Huang, Holger Hermanns, and Lijun Zhang.
\newblock Eidos: Efficient, imperceptible adversarial 3d point clouds.
\newblock \emph{arXiv preprint arXiv:2405.14210}, 2024{\natexlab{a}}.

\bibitem[Zhang et~al.(2023)Zhang, Chen, Liu, Ouyang, Xie, Zhu, Li, and
  Meng]{zhang2023meshattack}
Jinlai Zhang, Lyujie Chen, Binbin Liu, Bo Ouyang, Qizhi Xie, Jihong Zhu,
  Weiming Li, and Yanmei Meng.
\newblock 3d adversarial attacks beyond point cloud.
\newblock \emph{Information Sciences}, 633:\penalty0 491--503, 2023.

\bibitem[Zhang et~al.(2024{\natexlab{b}})Zhang, Dong, Zhu, Zhu, Kuang, and
  Yuan]{zhang2024improving}
Jinlai Zhang, Yinpeng Dong, Jun Zhu, Jihong Zhu, Minchi Kuang, and Xiaming
  Yuan.
\newblock Improving transferability of 3d adversarial attacks with scale and
  shear transformations.
\newblock \emph{Information Sciences}, 662:\penalty0 120245,
  2024{\natexlab{b}}.

\bibitem[Zhang et~al.(2022)Zhang, Zhang, Li, Song, and Gao]{zhang2022practical}
Qilong Zhang, Chaoning Zhang, Chaoqun Li, Jingkuan Song, and Lianli Gao.
\newblock Practical no-box adversarial attacks with training-free hybrid image
  transformation.
\newblock \emph{arXiv preprint arXiv:2203.04607}, 2022.

\bibitem[Zhang et~al.(2019)Zhang, Liang, Salem, and Jacobs]{zhang2019defense}
Yu Zhang, Gongbo Liang, Tawfiq Salem, and Nathan Jacobs.
\newblock Defense-pointnet: Protecting pointnet against adversarial attacks.
\newblock In \emph{2019 IEEE International Conference on Big Data (Big Data)},
  pages 5654--5660, 2019.

\bibitem[Zheng et~al.(2019)Zheng, Chen, Yuan, Li, and Ren]{zheng2019saliency}
Tianhang Zheng, Changyou Chen, Junsong Yuan, Bo Li, and Kui Ren.
\newblock Pointcloud saliency maps.
\newblock In \emph{Proceedings of the IEEE/CVF International Conference on
  Computer Vision}, pages 1598--1606, 2019.

\bibitem[Zhou et~al.(2019)Zhou, Chen, Zhang, Fang, Zhou, and Yu]{zhou2019dup}
Hang Zhou, Kejiang Chen, Weiming Zhang, Han Fang, Wenbo Zhou, and Nenghai Yu.
\newblock Dup-net: Denoiser and upsampler network for 3d adversarial point
  clouds defense.
\newblock In \emph{Proceedings of the IEEE/CVF international conference on
  computer vision}, pages 1961--1970, 2019.

\end{thebibliography}
}

\end{document}